\documentclass[10pt,twocolumn,letterpaper]{article}

\usepackage{iccv}              

\definecolor{iccvblue}{rgb}{0.21,0.49,0.74}
\usepackage[pagebackref,breaklinks,colorlinks,allcolors=iccvblue]{hyperref}

\def\paperID{9084} 
\def\confName{ICCV}
\def\confYear{2025}

\usepackage{colortbl}
\usepackage{multirow}
\usepackage{makecell}

\usepackage{float}
\usepackage{pifont}
\usepackage{xspace}

\definecolor{darkpink}{rgb}{0.8, 0.1, 0.4}

\usepackage{amsthm}

\newcommand{\method}{UAP\xspace}
\newcommand{\prob}{S-FDG\xspace}
\newcommand{\bC}[1]{\boldsymbol{\mathcal{#1}}}

\newcommand{\Cc}{\mathcal{C}}

\newcommand{\Lc}{\mathcal{L}}

\newcommand{\Nc}{\mathcal{N}}

\newcommand\pbf{\mathbf{p}}

\newcommand\qbf{\mathbf{q}}

\newcommand\wbf{\mathbf{w}}
\newcommand\xbf{\mathbf{x}}
\newcommand\ybf{\mathbf{y}}
\newcommand\zbf{\mathbf{z}}

\newcommand\Fbf{\mathbf{F}}
\newcommand\Gbf{\mathbf{G}}

\newcommand\Ibf{\mathbf{I}}

\newcommand\Sigmab{\boldsymbol{\Sigma}}

\newcommand\thetab{\boldsymbol{\theta}}

\newcommand\mub{\boldsymbol{\mu}}

\newcommand{\cmark}{\ding{51}}%
\newcommand{\xmark}{\ding{55}}%
\definecolor{darkgreen}{rgb}{0.7, 0.0, 0.0} 

\title{Unified Alignment Protocol: \\ Making Sense of the Unlabeled Data in New Domains}

\author{Sabbir Ahmed$^{1}$, Mamshad Nayeem Rizve$^{2}$, Abdullah Al Arafat$^{3}$, Jacqueline Liu$^{1}$, Rahim Hossain$^{1}$,\\
Mohaiminul Al Nahian$^{1}$, Adnan Siraj Rakin$^{1}$\\
$^{1}$Binghamton University (SUNY), $^{2}$Adobe Inc., $^{3}$North Carolina State University}

\begin{document}
\maketitle
\begin{abstract}
    Semi-Supervised Federated Learning (SSFL) is gaining popularity over conventional Federated Learning in many real-world applications. Due to the practical limitation of 
    limited labeled data on the client side, SSFL considers that participating clients train with unlabeled data, and only the central server has the necessary resources to access limited labeled data, making it an ideal fit for real-world applications (e.g., healthcare). However, traditional SSFL assumes that the data distributions in the training phase and testing phase are the same. In practice, however, domain shifts frequently occur, making it essential for SSFL to incorporate generalization  capabilities and enhance their practicality.
 The core challenge is improving model generalization to new, unseen domains while the client participate in SSFL. However, the decentralized setup of SSFL and unsupervised client training necessitates innovation to achieve improved generalization across domains. To achieve this, we propose a novel framework called the Unified Alignment Protocol (UAP), which consists of an alternating two-stage training process. The first stage involves training the server model to learn and align the features with a parametric distribution, which is subsequently communicated to clients without additional communication overhead. The second stage proposes a novel training algorithm that utilizes the server feature distribution to align client features accordingly. Our extensive experiments on standard domain generalization benchmark datasets across multiple model architectures reveal that proposed UAP successfully achieves SOTA generalization performance in SSFL setting.  
\vspace{-1.5em}
\end{abstract}

\begin{figure*}[t]
\begin{center}
\begin{tabular}{c}
\includegraphics [width=0.65\linewidth]{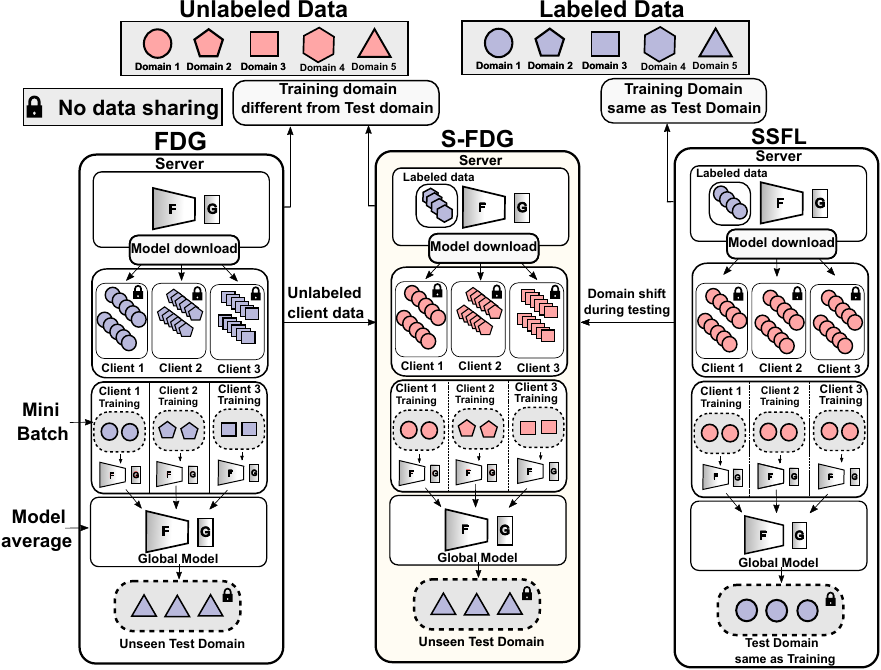}
 \end{tabular} 
 \vspace{-0.5em}
\caption{\emph{A comparative illustration of Federated Domain Generalization (FDG)~\cite{FedGMA,FedDG,FedDG_GA,fedsr}, Semi-Supervised Federated Learning (SSFL)~\cite{jeong2020federated,9671693,semiFL}, and our proposed Semi-Supervised Federated Domain Generalization (\prob). FDG (left) assumes clients have labeled data, with domain shift occurring during testing. SSFL (right) assumes clients have unlabeled data, the server has limited labeled data, but training and testing occur within a single domain. In contrast, \prob (middle) models a more realistic scenario where clients have unlabeled data, the server has limited labeled data, and domain shift occurs during testing.}}
\label{fig:ssfl-vs-uap}
\end{center}
\vspace{-2.7em}
\end{figure*}    
\section{Introduction}
\label{sec:intro}

Semi-Supervised Federated Learning (SSFL)~\cite{jeong2020federated,9671693,semiFL,semi_sup_fl_1,semi_sup_fl_2,semi_sup_fl_3} is becoming increasingly popular as a solution to overcome the challenge of obtaining labeled data at the client level. In numerous real-world applications~\cite{jeong2020federated}, clients may lack the resources, expertise, or inclination to engage in thorough data annotation. This is particularly true in on-device FL settings where specialized domain knowledge is often required for accurate labeling~\cite{9671693}. To resolve this, SSFL (shown in Figure~\ref{fig:ssfl-vs-uap}) offers a promising solution that operates under the practical assumption that while the clients have unlabeled data, the server possesses a limited amount of labeled data. In practice, the server is a powerful cloud-based service~\cite{server_powerful1,semiFL}
and often have access to some training data~\cite{server_dataset1, jeong2020federated, semiFL}.


These practical considerations make SSFL well-suited for real-world applications. For example, a speech-to-text conversion app aims to develop a robust model for transcribing audio while preserving consumer data privacy~\cite{speech_1,speech_2}. Here, the central server has access to a small labeled dataset, while the clients participate with unlabeled data. Another practical SSFL example would be low-resource clinical institutes (clients with unlabeled data) from various countries collaborating with a central high-resource clinical institute to construct a COVID-19 predictive model~\cite{naz2022comprehensive}. 
However, in both scenarios, maintaining consistent data distributions between the training and testing phases is challenging. For instance, a trained speech-to-text model may be deployed in regions with diverse accents and intonations, leading to domain shifts. This introduces the well-known Domain Generalization (DG) problem~\cite{muandet2013domain} into the SSFL framework (see Figure~\ref{fig:ssfl-vs-uap}), which we define as \emph{\textbf{S}emi-supervised \textbf{F}ederated \textbf{D}omain \textbf{G}eneralization (\prob)}. Our work addresses \prob by improving model generalization performance across unseen domains. 

Our preliminary investigation 
reveals that existing SSFL techniques~\cite{jeong2020federated,9671693,semiFL} do not generalize effectively across unseen domains primarily due to the additional constraint of domain shift. A potential solution would be to apply existing Domain Generalization (DG) techniques~\cite{ganin2016domain,sun2016deep,nguyen2021domain,nguyen2021kl,Li_2018_CVPR} in SSFL. However, these DG techniques are not designed for federated learning. We cannot directly adopt them in a privacy-preserving decentralized setup where data from servers and clients cannot be shared, whereas sharing multiple domain data is a common practice in DG~\cite{ganin2016domain,sun2016deep,nguyen2021domain,nguyen2021kl,Li_2018_CVPR}. Nevertheless, a related research direction known as Federated Domain Generalization (FDG)~\cite{FedDG,FedGMA,fedsr,FedDG_GA} (see Figure~\ref{fig:ssfl-vs-uap}) has tried to adopt and develop DG techniques for this decentralized setup. Again, our attempt to solve the \prob challenges via directly adopting FDG techniques underperforms (see Table~\ref{tab:challenges-ssfl}) primarily due to heavy dependence on labeled data on the client end. In summary, the problem of achieving \prob 
is under-studied, and existing methods in neither SSFL~\cite{jeong2020federated,9671693,semiFL}
 nor FDG~\cite{liu2021feddg,zhang2021federated,fedsr} adequately address this problem. \emph{In this paper, we are the first to tackle this research gap of achieving \prob and develop a novel training framework to solve it}.

 To address the challenges of \prob, our working principle is to learn domain invariant features across the server and client domains. However, learning domain invariant features faces two major roadblocks unique to \prob. First, the decentralized setup of SSFL prohibits individual data sharing between the server and clients, making it impossible to access multi-domain data simultaneously. Second, the constraint of unlabeled training data in clients make it even more challenging. To overcome these challenges, we propose a novel training method called \emph{Unified Alignment Protocol (\method)}. The proposed \method consists of alternating two-stage training, each designed to align the features between the client and server domains effectively. The first training stage, called \emph{Server Feature Alignment}, attempts to learn and align the server feature distribution with a standard parametric distribution (e.g., Gaussian) which the client can utilize in the next stage.
To make the server distribution accessible to the client training stage, we also ensure that this distribution parameters are communicated without additional communication overhead by embedding them into the model weight statistics. Next, the second training stage, called \emph{Client Feature Alignment}, aligns the individual client domain feature to the known server domain distribution communicated from the prior stage. For each stage, we develop novel training algorithm with loss function uniquely designed to learn a parametric distribution for the server and then align the client features accordingly. 


Following standard practice in the literature, we have evaluated the proposed \method across five Domain Generalization benchmark datasets and different model architectures.
Our proposed method consistently improves performance in \prob across
diverse datasets and architectures. In particular, the proposed \method can improve the generalization capabilities by improving the test domain accuracy by $\sim$37\% in an unseen test domain on the PACS dataset compared to the SOTA SSFL method~\cite{semiFL} without incurring additional communication overhead.

\vspace{-1em}
\section{Related Work}
\label{sec:related_work}
\vspace{-0.5em}

\noindent{\bf Semi-Supervised Federated Learning (SSFL). }
SSFL is a practical framework that assumes that labeled data exists only at the server end while clients have fully unlabeled data. The very first work that investigated this problem is FedMatch~\cite{jeong2020federated}. FedMatch~\cite{jeong2020federated} proposes the fusion of FL and Semi-Supervised Learning (SSL) methodologies by introducing an inter-client consistency loss. 
Building upon this,~\cite{9671693} refines FedMatch, establishing a robust baseline by mitigating gradient diversity.
The current SOTA method for SSFL is called SemiFL~\cite{semiFL}, which employs an \emph{alternate training} that trains server and clients in an alternating fashion. However, the major limitation of these methods is that they are designed to improve performance within the confines of a single domain, ignoring the critical issue of domain shift, which is prevalent in real-world applications.  In this work, we address this critical issue, aiming to improve \prob performance.
To the best of our knowledge, no existing work has explored the problem of \prob.
\noindent{\bf Federated Domain Generalization (FDG).} FDG~\cite{fedsr,FedDG,FedGMA,FedDG_GA} is one emerging research area that improves the generalization ability of trained models on unseen test domains while fully preserving FL's decentralized and privacy-preserving aspect. However, only a few studies have been carried out in this area, and most of them suffer from major drawbacks that limit their applications in real-world scenarios. For example,~\cite{FedDG} uses the amplitude spectrum on the frequency domain as the data distribution information and exchanges them among clients. However, the exchange operations introduce additional costs and increase risks of data privacy leakage. Meanwhile,~\cite{zhang2021federated} aligns the representation distribution across domains via a reference distribution from a generative model. While this fully preserves data privacy, it can be overly complicated to implement in practice. In comparison,~\cite{fedsr,FedGMA,FedDG_GA} offers viable solution that retains simplicity. However, even these methods are highly dependent on labeled data for generalization performance and hence proves to be ineffective for \prob (see Table~\ref{tab:comparison-with-ssfl}).

In summary, none of the existing SSFL and FDG approaches address the unique and under-explored paradigm of \prob.
Given the importance and practicality of \prob, we aim to develop a generalized SSFL framework.

\vspace{-1em}

\section{Problem Statement}
\vspace{-0.5em}
In this section, we aim to outline challenges of \prob and evaluate why existing methods fail to address them. 
\vspace{-0.5em}

\subsection{Preliminaries}


We use $\Gbf\circ\Fbf$ to denote the entire model, where $\Fbf(\cdot)$ is the feature extractor, and $\Gbf(\cdot)$ is the linear classifier. In addition, we use $\zbf = \Fbf(\xbf) \in \mathbb{R}^m$ to represent the $m$-dimensional feature vectors. The weights of the linear classifier $\Gbf(\cdot)$ are denoted by $\wbf_G \in \mathbb{R}^{m \times K}$, where $K$ represents the total number of classes in the class set $\Cc = \{1, \cdots, K\}$. The weight vector corresponding to the $k$-th class in $\wbf_G$ is represented as $\wbf_G^k \in \mathbb{R}^m$.




\noindent \textbf{SSFL.} Consider a scenario where there is a collaboration between a server and $M$ clients. 
And assume that the server has a
labeled dataset $\bC{D}_s = \{(\xbf_s^n, \ybf_s^n)\}_{n=1}^{N_s}$ with distribution $\pbf_s(\xbf,\ybf)$ whereas
each client $i \in [1, M]$ has an unlabeled dataset $\bC{D}_i = \{(\xbf_i^n)\}_{n=1}^{N_i}$ with distribution $\pbf_i(\xbf,\ybf)$. SSFL is an alternating two-stage training process that repeats over multiple communication rounds. First, the server utilizes its labeled dataset to train an initial model, which is then sent to the clients. The clients afterward utilize this model to produce pseudo-labels ($\Tilde{\ybf}$) for their unlabeled data and train their models. The clients then send their updated model to the server where the model aggregation occurs. 
In SSFL, the server and clients collaboratively aim to minimize the following objective function:
\begin{align}\label{eq:obj-function}
    \min_{\thetab} \ \bC{E}_s(\thetab) + \frac{1}{M} \sum_{i=1}^{M} \bC{E}_i(\thetab)
\end{align}
Here, $\thetab$ represents global model parameters. The server's objective function $\bC{E}_s(\thetab)$ is defined as:
\begin{align*}
    \bC{E}_s(\thetab) = \mathbb{E}_{\pbf_s(\xbf,\ybf)}[\bC{L}(\thetab; \xbf, \ybf)] \approx \frac{1}{N_s} \sum_{n=1}^{N_s} \bC{L}(\thetab; \xbf^{(n)}_s, \ybf^{(n)}_s),
\end{align*}
where $\bC{L}(\thetab; \xbf, \ybf)$ denotes the loss function for a data point $(\xbf, \ybf)$. And the client's local objective function $\bC{E}_i(\thetab)$ is defined as:
\begin{align*}
    \bC{E}_i(\thetab) = \mathbb{E}_{\pbf_i(\xbf,\ybf)}[\bC{L}(\thetab; \xbf, \hat{\ybf})] \approx \frac{1}{N_i} \sum_{n=1}^{N_i} \bC{L}(\thetab; \xbf^{(n)}_i, \Tilde{\ybf}^{(n)}_i).
\end{align*}

\noindent \textbf{\prob.} The objective in \prob is to train a global model that can effectively generalize to an unseen test domain $\pbf_T(\xbf, \ybf) \sim \bC{T}$, where $\bC{T}$ represents a family of distribution. Here, the goal is not only to minimize the expected loss in seen domains but also to minimize the loss on unknown test domains, i.e.,
\begin{align*}
\min_{\thetab} \ \mathbb{E}_{\pbf_T \sim \bC{T}} \left[ \mathbb{E}_{\pbf_T( \xbf,\ybf)}[\bC{L}(\thetab; \xbf, \ybf)] \right]    
\end{align*}

\subsection{Challenges of \prob}\label{sec:ssfl-challenges}
%
Existing SSFL approaches assume that training and test data come from a single domain--an assumption that does not often hold in real-world scenarios. In real-world scenarios, it is highly likely to encounter a situation where training and test data have data from different domains, i.e.,
\begin{align*}
    & \pbf_s(\xbf, \ybf) \neq \pbf_T(\xbf, \ybf), \\
    & \pbf_i(\xbf, \ybf) \neq \pbf_T(\xbf, \ybf), \quad \forall i
\end{align*}
For instance, consider a scenario where multiple low-resource clinical institutes (training clients with unlabeled data) from different countries collaborate with a central high-resource clinical institute (server with labeled data) to develop a predictive model for COVID-19~\cite{dayan2021federated}. The aim is not merely to perform well on local data of participating countries (training data) but also to perform effectively on data from other countries (test data with domain shift). 

\noindent\textbf{Observations.} After applying the SOTA SSFL method~\cite{semiFL} to handle this realistic setting, we observe that it performs poorly to generalize to unseen test domains (e.g., $24.95\%$ accuracy), as evident from Table~\ref{tab:challenges-ssfl}. This leads us to our first observation:



\begin{table}[t]
\centering\setlength{\tabcolsep}{5pt}
\caption{\emph{DG performance of SOTA SSFL, FDG and \method on PACS dataset. The table displays results across three server training domains: Cartoon, Photo, Sketch and reports test performance on unseen Art Painting domain (see Experimental Setup Section).}}
\scalebox{0.7}{
\begin{tabular}{lcccccc}
\toprule
Methods & Single  & Centralized  & Client & Cartoon & Photo & Sketch \\
& Domain & Setup & Labels & & & \\
\midrule
SOTA SSFL~\cite{semiFL} & \cmark & \xmark & \xmark & 52.20 & 52.39 &  24.95 \\
SOTA FDG~\cite{FedDG_GA} & \xmark & \xmark & \cmark & 24.46 & 18.50 & 17.72 \\
\rowcolor[gray]{.95} \method (Ours) & \xmark & \xmark & \xmark &  \textbf{75.73} & \textbf{64.84} & \textbf{61.67}\\
\bottomrule
\end{tabular}}  
\label{tab:challenges-ssfl}
\vspace{-1em}
\end{table}


\medskip

\textbf{Observation I.} {\em Existing SSFL approach do not generalize well to unseen test domains.}

\medskip


Next, to address the \prob problem, an alternative would be incorporating existing FDG methods~\cite{FedDG,FedGMA,FedDG_GA}. Since the domain generalization problem inspires FDG methods, they have adopted and developed new techniques to resolve the DG problem in federated learning. One challenge in applying FDG in a \prob setting is the lack of available label data. Nevertheless, we adopted the approach in~\cite{shot} to generate pseudo labels on the client side and directly apply FDG techniques to solve the \prob problem.
However, extending existing FDG methods has proven ineffective (with low accuracy) in addressing \prob as shown in Table~\ref{tab:challenges-ssfl}. This leads us to our second observation:


\medskip

\textbf{Observation II.} {\em Existing FDG techniques underperforms to achieve \prob.}


\medskip

In summary, the dual challenge of decentralized training setup and unlabeled client data makes achieving \prob daunting. Hence, motivated by the practicality of the frequently encountered domain shift problem, we propose a novel \prob training approach to address this challenging problem for the first time. 
Our method's principle is that since the clients cannot share their data, the possible solution is to learn domain-invariant features across server and client domains. 

\vspace{-1em}
\section{Our proposed \method Method}
\vspace{-0.5em}
We propose a novel training method called \emph{Unified Alignment Protocol (\method)}, which aims to learn domain invariant features given this challenging decentralized setting of SSFL with unlabeled client data. 
Our proposed \method is a novel alternating two-stage training process that continues over multiple communication rounds. The first training stage is called \emph{Server Feature Alignment}, where we propose a novel training algorithm to learn and align the server feature distribution with a standard parametric distribution (e.g., Gaussian). This stage aims to align the server feature distribution to a standard parametric distribution which the client can utilize in the next stage. The server then communicate these distribution parameters embedded into model weight statistics to the client without additional communication overhead. In the second training stage, called \emph{Client Feature Alignment Stage}, we develop a novel client training step to align client features with server features communicated from the prior stage. In summary, our proposed \method
aligns the features of multiple clients and servers through three strategies: i) train the server model using labeled data to learn and align with a parametric distribution (e.g., Gaussian), ii) communicate the information of this known distribution to the client efficiently, iii) train the client features to align with the server feature distribution without any labeled data.

\begin{figure*}[t]
\centering
\includegraphics[width=0.63\linewidth]{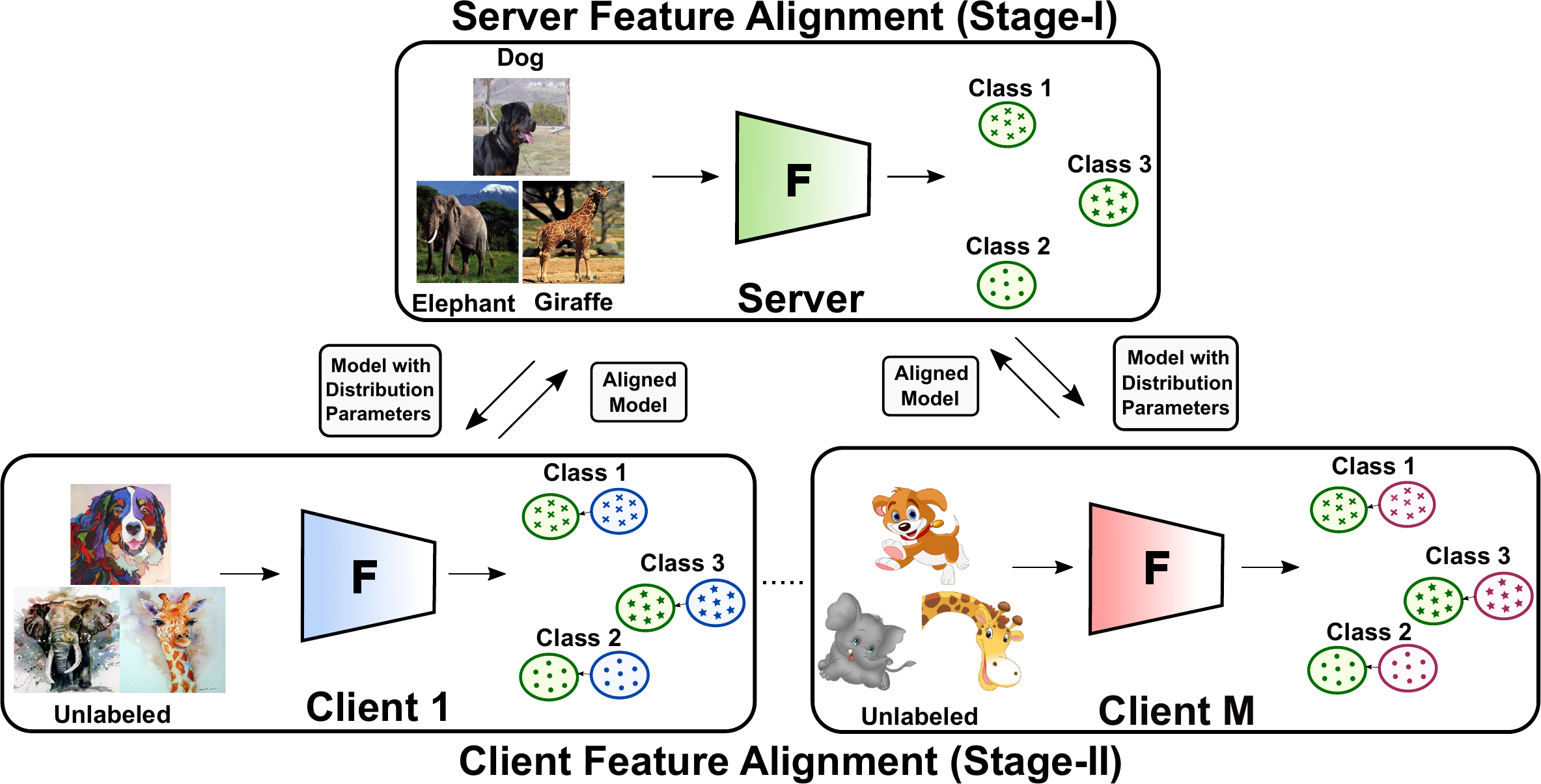}
\\
\caption{\emph{Overview of our proposed \method, where the server model is trained to learn and align feature with a parametric distribution (Sever Feature Alignment). Then, the server conveys both the model and its feature distribution parameters (no communication overhead) to the client by embedding them into the model parameters. Clients then leverage the server feature distribution knowledge to align their features (Client Feature Alignment) accordingly.}}
\vspace{-1.7em}
\end{figure*}

\subsection{Stage-I: Server Feature Alignment}
Our first training stage trains the server to learn and align features with a standard parametric distribution. The goal is to learn the server feature distribution parameters, which the client can leverage later to align their features. To achieve this, we train the server model by guiding the feature representations $\zbf$ per class
to follow Gaussian distribution:
\begin{align*}
    \pbf_s(\mathbf{z}|\mathbf{y}=k) = \Nc_k(\mub_k, \Sigmab_k), \quad \forall k \in \Cc
    \vspace{-0.5em}
\end{align*}
 We aim to communicate the feature distribution to the clients after server training so that the clients can align their features with the server. 

In our stage-I training, we introduce two novel strategies to align feature representations with a standard parametric distribution. First, we train the server model by guiding the classifier weight vector $\wbf_G^k$ to align with the mean of the feature distribution for the $k$-th class. This alignment ensures that, by the end of the server training stage, the classifier weights satisfy the following equality:
\begin{align*}
    \wbf_G^k = \mub_k, \quad \forall k \in \mathcal{C}
\end{align*}
 As a result, the distribution mean parameters $\mub_k$ of each class $k \in \Cc$ can be directly obtained from the server classifier weights. Second, we guide the covariance matrix \(\Sigmab_k\) of class features to take a diagonal form, specifically \(\lambda \Ibf\), where \(\lambda\) is a small positive constant. This strategy ensures that the distribution covariance parameters \(\Sigmab_k\) are explicitly known and simplifies the feature distribution structure. Additionally, by keeping \(\lambda\) small, we encourage features of each class to remain closely clustered around their respective means, promoting more discriminative representations.

 

\noindent\textbf{Communication overhead reduction.} In addition to providing knowledge of distribution parameters, our two strategies significantly enhance communication efficiency by reducing the cost of transmitting distribution parameters. Typically, to communicate \(\pbf_s(\zbf|\ybf)\), the server must transmit the mean \(\mub_k\) and covariance \(\Sigmab_k\) of the distribution \(\mathcal{N}_k(\mub_k, \Sigmab_k)\) for each class \(k \in \mathcal{C}\), leading to a substantial communication overhead of \(K \cdot m + K \cdot m^2\). Specifically, transmitting the mean \(\mub_k\) for each of the \(K\) classes incurs an additional overhead of \(K \cdot m\), as each class has an \(m\)-dimensional feature space. Furthermore, transmitting the covariance \(\Sigmab_k\) for each class results in an additional overhead of \(K \cdot m^2\).  To exacerbate the problem, the communication burden scales with both the number of classes \(K\) and the feature dimension \(m\), making it increasingly prohibitive. For instance, in a ResNet-50 model with 1000 classes, this overhead is approximately 176 times the size of the model's parameters, creating a significant communication overhead challenge. However, our proposed training strategies eliminate this overhead entirely. Since the class means \(\mub_k\) are directly obtained from the classifier weights, they no longer need to be transmitted. Additionally, we assume that participating clients in our proposed \prob setting are aware that the covariance matrices are diagonal, which removes the necessity of communicating \(\Sigmab_k\). Next, to incorporate these principles into our proposed \method training scheme, we develop a novel server training loss function outlined in the following subsection.

\noindent \textbf{Server Training Objective:} To achieve the above goals,  our proposed loss function for the server training stage is as follows
\begin{align*}
    \mathbf{\Lc_{server}} = \mathbf{\Lc_{CE}}(\hat{\ybf}, \ybf) + \alpha \cdot \mathbf{\Lc_{CDD}} + \beta \cdot \mathbf{\Lc_{COV}}
\end{align*}
where $\mathbf{\Lc_{CE}}$ represents standard cross-entropy loss for classification, $\mathbf{\Lc_{CDD}}$ is the Contrastive Domain Discrepancy loss and $\mathbf{\Lc_{COV}}$ is the covariance matching loss and $\alpha$ and $\beta$ are hyperparameters used to couple the loss terms. Next, we outline the role of each proposed loss component of our method i.e., $\mathbf{\Lc_{CDD}}$ \& $\mathbf{\Lc_{COV}}$.


\noindent \textbf{Contrastive Domain Discrepancy Loss ($\mathbf{\Lc_{CDD}}$).} Next, we want to learn features $\zbf$ where the feature distribution per class is multivariate Gaussian. We choose Gaussian distribution because it is standard practice to assume Gaussian distribution for intra-class features~\cite{why_gaussian1,why_gaussian2,why_gaussian3}. To achieve this, we create $K$ dynamic Gaussian distributions as follows
\vspace{-0.2em}
\begin{align}\label{eq:dyn-gauss-dist}
    \qbf_k = \Nc_k(\wbf_G^k, \lambda \cdot \Ibf), \quad \forall k \in \Cc
\end{align}
where the mean of the distributions are set using the classifier weight matrix $\wbf_G$, and the covariance matrices are diagonal in structure. The scaling factor, $\lambda$, used in this process is a hyperparameter that is independent of class $k$. Our proposed method guides the features $\zbf = \Fbf(\xbf)$ for each class $k \in \Cc$ to align with the dynamic Gaussian distribution $\qbf_k$ evolving with server training. 

To achieve this alignment, we propose to utilize Contrastive Domain Discrepancy (CDD) loss~\cite{kang2019contrastive}. First, we randomly select a subset $\Cc' \subset \Cc$ to compute this loss. For every $k \in \Cc'$, we select $N_k$ images from $\bC{D}_{s}$ which we pass through the feature extractor $\Fbf(\cdot)$ to get mini-batch of server features $\{\zbf_i^k\}_{i=1}^{N_k}$. In addition, we also sample $N_k$ features $\{\Tilde{\zbf}_j^{k}\}_{j=1}^{N_k}$ from our dynamic Gaussian distribution $\qbf_k$ defined in Eq.~\eqref{eq:dyn-gauss-dist}
to create a targeted Gaussian feature mini-batch. Then, using these mini-batches, we compute a class-conditioned version of Maximum Mean Discrepancy (MMD) to measure the difference between the server feature distribution $\pbf_s(\zbf|\ybf=k_1)$ and dynamic Gaussian distribution $\qbf_{k_2}$ for each pair of classes $k_1, k_2 \in \Cc'$,
\begin{align*}
\bC{L}_{k_1,k_2}^{\textbf{MMD}} &= \sum_{i=1}^{N_{k_1}} \sum_{j=1}^{N_{k_1}} \frac{\mathbb{K}(\zbf_{i}^ {k_1}, \zbf_{j}^ {k_1})}{N_{k_1}^2} + \sum_{i=1}^{N_{k_2}} \sum_{j=1}^{N_{k_2}} \frac{\mathbb{K}(\Tilde{\zbf}_{i}^ {k_2}, \Tilde{\zbf}_{j}^ {k_2})}{N_{k_2}^2} \\ &-2 \sum_{i=1}^{N_{k_1}} \sum_{j=1}^{N_{k_2}} \frac{\mathbb{K}(\zbf_{i}^ {k_1}, \Tilde{\zbf}_{j}^ {k_2})}{N_{k_1} \cdot N_{k_2}}
\end{align*}
where $\mathbb{K}(\cdot, \cdot)$ is a kernel function that embeds feature representations into a Reproducing Kernel Hilbert Space (RKHS). Finally, utilizing this loss, we compute the CDD loss as follows
\begin{align}\label{eq:cdd-loss}
\bC{L}_{\textbf{CDD}} = \frac{\sum\limits_{k \in \Cc'} \bC{L}_{k,k}^{\textbf{MMD}}}{|\Cc'|} - \frac{\sum\limits_{k_1 \in \Cc'} \sum\limits_{k_2 \in \Cc'}^{k_1 \neq k_2} \bC{L}_{k_1,k_2}^{\textbf{MMD}}}{|\Cc'|(|\Cc'| - 1)}
\end{align}
Here, the first term represents intra-class discrepancy between $\pbf_s(\zbf|\ybf=k)$ and $\qbf_k$ to be diminished, and the second term represents interclass discrepancy between $\pbf_s(\zbf|\ybf=k_1)$ and $\qbf_{k_2}$ for $k_1 \neq k_2$ to be enlarged. In summary, the first term in CDD loss facilitates the alignment, and the second term ensures that the features are highly discriminative.

\noindent \textbf{Covariance Matching Loss ($\mathbf{\Lc_{COV}}$).} Finally, we add a covariance regularization term that implicitly promotes covariance matching between the server and the clients. This will help the client overcome the challenges of matching with server distribution later without the label. 
We explain the rationale behind this regularization in more detail in the following Subsection. Our covariance regularization term is defined as
\begin{align}\label{eq:cov-reg}
    \bC{L}_{\textbf{COV}} = \frac{1}{m}||\Sigmab_{z} - \Sigmab||_2^2
\end{align}
where $\Sigmab_{z}$ is the covariance of features calculated for each minibatch and $\Sigmab$ is a reference covariance matrix detailed in following subsection.
After server training is completed, the server communicates to the clients that the per-class feature distribution on the server side follows Gaussian. Additionally, the server conveys that the classifier weights serve as the means, and the covariance matrices have a diagonal structure. With this information, clients gain sufficient knowledge about the server feature distribution $\pbf_s(\zbf|\ybf)$ without additional communication overhead.

\subsection{Stage-II: Client Feature Alignment}\label{sec:alignment-stage}

In this stage, we aim to train the client models to align the client features such that the $i$-th client distribution $\pbf_i(\zbf|\ybf)$ matches with server distribution $\pbf_s(\zbf|\ybf)$. In SSFL, the clients do not have labels for their data. This introduces an additional challenge to match $\pbf_s(\zbf|\ybf)$ with $\pbf_i(\zbf|\ybf)$ since $\ybf$ is unknown at client end. To resolve this issue, we generate pseudo labels $\Tilde{\ybf}$ and match client distribution $\pbf_i(\zbf|\Tilde{\ybf})$ with server distribution $\pbf_s(\zbf|\ybf)$. However, prior works~\cite{wang2021gpl,WANG2023108987} reveal that pseudo-label quality degrades with domain shift. Hence, to mitigate the shortcomings of noisy pseudo labels, we propose a novel label-independent covariance regularization loss that improves domain invariant feature learning.

While it is a common practice~\cite{coral, covariance_matching2,covariance_matching3} to learn domain invariant features by minimizing the difference between covariance matrices of features from different domains, the decentralized setting of SSFL makes it challenging to align server and client covariance matrices directly by sharing data. Our proposed covariance regularization loss is designed to address this problem. We apply the covariance regularization loss during both server and client training, which guides the server and client feature covariance matrices to align with a reference matrix. As demonstrated in~\cite{cogswell2015reducing}, diagonal covariance matrices, which decorrelate features, exhibit superior generalization properties while mitigating overfitting. Motivated by this, we set the reference matrix as a diagonal matrix ($\Sigmab = \gamma \Sigmab_k$), where $\gamma$ is a hyperparameter. To further investigate the impact of this choice, we conduct an ablation study on different reference matrices (see Supplementary). Thus, our proposed covariance regularization implicitly aligns the server and client covariance matrices. By minimizing the difference between the server and client domain's feature covariance matrix, we expect to reduce the problem of noisy pseudo-labels due to decentralized multi-domain clients~\cite{covariance_matching2}.  



\noindent \textbf{Client Training Objective:} First, we generate pseudo labels for clients' unlabeled datasets using the server model. To compute pseudo labels, we first compute class centroids for the client data using weighted k-means clustering and assign pseudo labels based on the nearest class centroid~\cite{shot}. We then train the client model to minimize standard cross-entropy loss calculated between the predicted label $\hat{\ybf}$ and the pseudo label $\Tilde{\ybf}$. Additionally, to align features between clients and servers, first, we generate the server feature distributions per class using 
the classifier weight $\wbf_G$  as follows
\begin{align*}
    \pbf_s(\zbf|\ybf = k) = \Nc_k(\wbf_G^k, \lambda \cdot \Ibf), \quad \forall k \in \Cc 
\end{align*}
Then, we use $\bC{L}_{\textbf{CDD}}$ loss defined in Eq.~\eqref{eq:cdd-loss} to align the client feature distribution $\pbf_i(\zbf|\Tilde{\ybf})$ with server feature distribution $\pbf_s(\zbf|\ybf)$ during client training. 

Finally, we add the covariance regularization term defined in Eq.~\eqref{eq:cov-reg} on the client-side training to match the covariance of the server. Thus, the overall loss function for training the client is as follows
\begin{align*}
    \mathbf{\Lc_{client}} = \mathbf{\Lc_{CE}}(\hat{\ybf}, \Tilde{\ybf}) + \alpha \cdot \mathbf{\Lc_{CDD}} + \beta \cdot \mathbf{\Lc_{COV}}
\end{align*}
where $\mathbf{\Lc_{CE}}$ represents cross-entropy loss, $\mathbf{\Lc_{CDD}}$ is the Contrastive Domain Discrepancy loss and $\mathbf{\Lc_{COV}}$ is the covariance matching loss and $\alpha$ and $\beta$ are hyperparameters used to couple the loss terms. 

Finally, both server and client stages are repeated alternatively for multiple communication rounds to develop a novel \prob training framework.

\vspace{-0.5em}

\section{Experimental Setup}\label{sec:exp_setup}

\noindent \textbf{Datasets and Models.}
We evaluate our method extensively using five established DG benchmark datasets: PACS~\cite{yu2022pacs} (four domains), VLCS~\cite{6751316} (four domains), OfficeHome~\cite{venkateswara2017deep} (four domains), TerraIncognita~\cite{terra_incognita} (four domains) and RotatedMNIST~\cite{qiao2023scale} (six domains). For additional details of the dataset, we direct the reader to the Supplementary section. To evaluate performance on the PACS, VLCS, OfficeHome and TerraIncognita datasets, we use the ResNet18 model, and for the RotatedMNIST dataset, we utilize the network architecture detailed in~\cite{fedsr}. We also evaluate our method's performance with other model architectures (see Table~\ref{tab:different-models}).

\noindent \textbf{Evaluation Metrics and Hyperparameters.} For performance evaluation, we allocated one domain as the server dataset and another as the unseen test domain for evaluating the final global model, while the remaining domains were assigned to individual clients. Specifically, in a dataset with $N_d$ domains, one domain is used for the server, another for testing, and the remaining $N_d-2$ domains are distributed among $N_d-2$ clients. This setup is consistent with existing FDG methods~\cite{fedsr,FDG_GA,stablefdg,feddg_cvpr_2024}, where each client is provided with data from a unique domain. The accuracy of the final global model is reported on the unseen test domain. Additionally, we assess performance by increasing the number of clients by splitting a single domain's data among multiple clients (See Sec~\ref{sec:ablation}). To train both the server and client models, we set the batch size to 64 and initialized the learning rate at 0.002. We used SGD as the optimizer, with a cosine learning rate decay as the scheduler. The number of local epochs was set to 5, and the total number of communication rounds was 40. Additionally, we set  $\alpha=1$, $\beta=1$, $\gamma=100$ and $\lambda=0.01$ for all our experiments. We direct the reader to the Supplementary section for additional implementation details and ablation studies of hyper-parameters (e.g., $\alpha$, $\beta$, $\gamma$ and $\lambda$). 

\noindent \textbf{Baseline SSFL and SOTA Methods.} We compare our proposed \method with a baseline SSFL setting~\cite{semiFL}, where the server is trained using labeled data, and the clients are trained using pseudo labels with standard cross-entropy loss. It is important to note that the experimental setup for the baseline SSFL is identical to that of the proposed \method, with the only difference being the loss function used. Since we are the first to address the \prob problem, there are no existing SOTA methods specifically targeting \prob for comparison. However, we report the results of the current SOTA SSFL and FDG techniques (see Table~\ref{tab:comparison-with-ssfl}) and compare them with our proposed approach.

\begin{table}[t]
\centering
\setlength{\tabcolsep}{5pt}
\caption{\em Performance comparison of baseline SSFL and \method across two test domains A and C from PACS dataset (rest are in supplementary).  The PACS dataset consists of four domains: Art Painting (A), Cartoon (C), Photo (P), and Sketch (S). For each combination, we allocated one domain as the server training dataset, another as the unseen test domain for the final global model while assigning the remaining domains to individual clients.}
\scalebox{0.67}{
\begin{tabular}{lc|ccc|ccc}
\toprule
\multirow{2}{*}{Method} & \multicolumn{1}{|c|}{Unseen Test Domain} & \multicolumn{3}{c|}{\textbf{A}} & \multicolumn{3}{c}{\textbf{C}} \\ 
\cline{2-8} 
& \multicolumn{1}{|c|}{Server Trained on} & C & P & S & A & P & S \\ \midrule
SSFL & & 64.65 & 40.19 & 36.28 & 69.97 & 21.20 & 37.12 \\ \hline
\rowcolor[gray]{0.95} \method (Ours) & & \textbf{75.73} & \textbf{64.40} & \textbf{67.92} & \textbf{70.65} & \textbf{51.96} & \textbf{67.49} \\ \bottomrule
\end{tabular}}
\label{tab:pacs_ac}
\vspace{-0.5em}
\end{table}

\begin{table}[t]
\centering\setlength{\tabcolsep}{5pt}
\caption{\em Performance comparison of baseline SSFL and UAP across two test domains C and V from VLCS dataset (rest are in supplementary). The VLCS dataset comprises four domains: Caltech101 (C), VOC2007 (V) LabelMe (L) and SUN09 (S). For each combination, we allocated one domain as the server training dataset, another as the unseen test domain for the final global model while assigning the remaining domains to individual clients. }
\scalebox{0.67}{
\begin{tabular}{lc|ccc|ccc}
  \toprule
  \multirow{2}{*}{Method} & \multicolumn{1}{|c|}{Unseen Test Domain}  & \multicolumn{3}{c|}{\textbf{C}} & \multicolumn{3}{c}{\textbf{V}} \\ 
  \cline{2-8} 
                     &  \multicolumn{1}{|c|}{Server Trained on}   & L & S & V & C & L & S \\ \midrule
  SSFL      &          & 47.28   & 14.77 & 94.35   & 47.04      & 38.30   & 49.29 \\ 
  \rowcolor[gray]{0.95} UAP (Ours) &         & \textbf{95.41}   & \textbf{64.31} & \textbf{97.88}   & \textbf{54.59}      & \textbf{66.91}   & \textbf{62.97} \\ \bottomrule
\end{tabular}} 
\label{tab:vlcs_cv}
\vspace{-0.5em}
\end{table}

\begin{table}[t]
\centering\setlength{\tabcolsep}{5pt}
\caption{\em Performance comparison of baseline SSFL and \method across two test domains A and C in the OfficeHome dataset (rest are in supplementary). The OfficeHome dataset consists of four domains: Art (A), Clipart (C), Product (P), and Real (R). For each combination, we allocated one domain as the server training dataset, another as the unseen test domain for the final global model while assigning the remaining domains to individual clients.}
\scalebox{0.67}{
\begin{tabular}{lc|ccc|ccc}
  \toprule
  \multirow{2}{*}{Method} & \multicolumn{1}{|c|}{Unseen Test Domain} & \multicolumn{3}{c|}{\textbf{A}} & \multicolumn{3}{c}{\textbf{C}} \\ 
  \cline{2-8} 
                         & \multicolumn{1}{|c|}{Server Trained on} & C & P & R & A & P & R \\ \midrule 
  SSFL               & & 44.38 & 40.75 & 53.23 & 39.06 & 39.04 & 46.92 \\ 
  \rowcolor[gray]{0.95} \method (Ours)     &     & \textbf{48.95} & \textbf{47.51} & \textbf{55.95} & \textbf{44.28} & \textbf{44.51} & \textbf{48.48} \\ \bottomrule
\end{tabular}}
\label{tab:office-home_ac}
\vspace{-1.3em}
\end{table}

\vspace{-0.5em}
\section{Experimental Results}

\subsection{Evaluation of \method}
The evaluation of \method on the PACS and VLCS datasets is presented in Table~\ref{tab:pacs_ac} and Table~\ref{tab:vlcs_cv}, respectively. The results demonstrate a consistent performance improvement over SSFL across the test domains on both datasets. For instance, on the PACS dataset, we observe an average accuracy improvement of over 22\% on the unseen test domain when Art Painting is selected, and a similar gain is noted on the Cartoon test domain. On the VLCS dataset, our method achieves an average accuracy improvement of more than 33\% when Caltech101 is the test domain and more than 16\% when VOC2007 is the test domain. The reason for the variability in performance can be attributed to the difficulty of each domain task, as some domains are difficult to generalize, resulting in a weaker feature alignment.


Nevertheless, \method consistently improves the generalization to test domain across different datasets, including OfficeHome dataset as shown in Table~\ref{tab:office-home_ac}. Again, on average, on OfficeHome dataset, proposed \method yields over $4\%$ accuracy improvement compared to SSFL across Art and Clipart test domains. The detailed results for the RotatedMNIST dataset and other test domains are provided in the supplementary materials. A general conclusion across different datasets is that our proposed \method can significantly improve DG performance across most test domains.

\begin{table}[t]
\centering
\centering\setlength{\tabcolsep}{5pt}
\caption{\em Comparative DG performance of SSFL and FDG methods trained with Pseudo labels and our proposed \method on PACS dataset. 
The GAIN column shows performance improvement of our method compared to the second best method (highlighted by underline).}
\scalebox{0.65}{
\begin{tabular}{lcccccc}
\toprule
Method &  Cartoon  & \color{darkgreen}{GAIN} &  Photo & \color{darkgreen}{GAIN} & Sketch & \color{darkgreen}{GAIN} \\
\midrule
CBAFed~\cite{cbafed} & 45.41 & & 48.49 & & 15.33 & \\
FedDG~\cite{FedDG} & 41.70 & & 34.52 & & \underline{52.15} & \\
FedDG-GA~\cite{FedDG_GA} & 24.46 & & 18.50 & & 17.72 & \\
FedGMA~\cite{FedGMA} & 41.11 & & 16.94 & & 27.93 & \\
FedSR~\cite{FedGMA} & 29.34 & & 19.94 & & 27.49 & \\
RScFed~\cite{rscfed} & \underline{65.91} & & 14.40 & & 31.20 & \\
SemiFL~\cite{semiFL} & 52.20 & & \underline{52.39} & & 24.95 & \\
\rowcolor[gray]{0.95} \method (Ours)  & \textbf{75.73} & \color{darkgreen}{+9.82} & \textbf{64.40} & \color{darkgreen}{+12.01} & \textbf{67.92} & \color{darkgreen}{+15.77} \\
\bottomrule
\end{tabular}}
\label{tab:comparison-with-ssfl}
\vspace{-1em}
\end{table}


\subsection{Comparison with SOTA SSFL/FDG}
\vspace*{-0.25em}

In Table~\ref{tab:comparison-with-ssfl}, we compare \method with SOTA SSFL methods~\cite{semiFL,rscfed,cbafed} as well as SOTA FDG methods~\cite{FedDG,fedsr,FedGMA,FedDG_GA}. We report performance of global model on the unseen Art Painting domain of PACS~\cite{yu2022pacs} dataset. More results on other datasets (OfficeHome and TerraIncognita) are reported in supplementary. To train using FDG methods, we pretrain the server model with the server dataset and generate pseudo labels prior to client training.
The results indicate that the current SSFL methods struggles with Domain Generalization (DG), especially evident in the Photo and Sketch server domains. However, the performance gap narrows when using the Cartoon domain as the server. Again, this discrepancy can be attributed to the similarity between the Art Painting and Cartoon domains compared to the large domain shifts between Art Painting and Photo or Art Painting and Sketch. On the other hand, the FDG approaches are incapable of enhancing DG performance, even when trained using pseudo labels since these methods rely heavily on client-labeled data. Nonetheless, whereas existing SSFL and FDG methods struggles, our method thrives on them and successfully generalizes across domains. These results successfully establish the significance of our proposed \method for achieving \prob.
\vspace{-0.5em}
\begin{table}[h]
\centering
\setlength{\tabcolsep}{5pt}
\caption{\em Effect of different loss components on \method evaluated on PACS dataset. The table displays the results across two server training domains: Cartoon, Photo with test performance of the global model reported on the unseen Art Painting domain.}
\scalebox{0.7}{
\begin{tabular}{lcc}
  \toprule
  Loss & Cartoon & Photo \\ \midrule
  $\bC{L}_{\textbf{CE}}$  & 64.65 & 40.19 \\ 
  $\bC{L}_{\textbf{CE}} + \alpha \cdot \bC{L}_{\textbf{CDD}}$ & 72.61 & 58.25 \\ 
  \rowcolor[gray]{0.95} $\bC{L}_{\textbf{CE}} + \alpha \cdot \bC{L}_{\textbf{CDD}} +  \beta \cdot \bC{L}_{\textbf{COV}}$ (\method) & \textbf{75.73} & \textbf{64.40} \\ \bottomrule
\end{tabular}}
\label{tab:loss-components}
\vspace{-0.5em}
\end{table}

\begin{table}[h]
\centering
\setlength{\tabcolsep}{5pt}
\caption{\em Performance with different model architectures on PACS dataset across two server training domains.}
\scalebox{0.7}{
\begin{tabular}{lcccccccc}
\toprule
Method & \multicolumn{2}{c}{VGG11} & \multicolumn{2}{c}{ResNet18} & \multicolumn{2}{c}{DenseNet121} & \multicolumn{2}{c}{DeiT-B} \\
\cmidrule(lr){2-3} \cmidrule(lr){4-5} \cmidrule(lr){6-7} \cmidrule(lr){8-9}
& Cartoon & Photo & Cartoon & Photo & Cartoon & Photo & Cartoon & Photo \\
\midrule
SSFL & 59.33 & 23.44 & 64.65 & 40.19 & 74.51 & 61.72 & 89.45 & 75.44 \\
\rowcolor[gray]{0.95} \method & \textbf{63.33} & \textbf{56.59} & \textbf{75.73} & \textbf{64.40} & \textbf{80.86} & \textbf{66.89} & \textbf{90.04} & \textbf{83.84} \\
\bottomrule
\end{tabular}}
\label{tab:different-models}
\vspace*{-0.5em}
\end{table}

\begin{table}[h]
\centering
\setlength{\tabcolsep}{5pt}
\caption{\em Performance with different number of clients (M) on PACS dataset across two server training domains.}
\scalebox{0.7}{
\begin{tabular}{lcccccc}
\toprule
Method & \multicolumn{2}{c}{M=2} & \multicolumn{2}{c}{M=6} & \multicolumn{2}{c}{M=8}\\
\cmidrule(lr){2-3} \cmidrule(lr){4-5} \cmidrule(lr){6-7}
& Cartoon & Photo & Cartoon & Photo & Cartoon & Photo \\
\midrule
SSFL & 64.65 & 40.19 & 61.08 & 32.47 & 70.31 & 31.20 \\
\rowcolor[gray]{0.95} \method & \textbf{75.73} & \textbf{64.40} & \textbf{73.34} & \textbf{63.14} & \textbf{75.05} & \textbf{64.21} \\
\bottomrule
\end{tabular}}
\label{tab:different-clients}
\vspace*{-0.5em}
\end{table}

\vspace{-0.7em}
\subsection{Ablation Study}\label{sec:ablation}
\vspace{-0.5em}
All our ablation studies are conducted using  PACS~\cite{yu2022pacs} benchmark dataset, with Art Painting as the unseen test domain and Cartoon and Photo as the server domains.

\noindent \textbf{Effect of different loss components.} Table~\ref{tab:loss-components} demonstrates the effect of each component of our proposed loss function in \method. Starting with the baseline performance using only \(\mathcal{L}_\mathbf{CE}\) loss, we observe a significant improvement with the addition of our alignment loss component \(\mathcal{L}_\mathbf{CDD}\). This highlights that \(\mathcal{L}_\mathbf{CDD}\) successfully facilitates in learning domain invariant features given a decentralized SSFL. Furthermore, the addition of covariance regularization loss \(\mathcal{L}_\mathbf{COV}\) helps reduce the challenges of noisy pseudo labels across multiple domains by increasing the performance by additional $\sim4\%$.

\noindent \textbf{Evaluation with different model architectures.} Table~\ref{tab:different-models} presents a comparison between SSFL and the proposed \method, conducted on three CNN architectures: VGG11, ResNet18, DenseNet121 and a Vision Transformer, DeiT-B. The results demonstrate that our proposed \method improves performance across a wide range of model architecture, further enforcing its adaptability and generalizability.

\noindent \textbf{Effect of number of clients.} Table~\ref{tab:different-clients} presents a comparison between SSFL and the proposed \method with different number of clients. The results demonstrate that our proposed \method improves performance compared to SSFL. However, performance decreases slightly with increasing number of clients due to more decentralization which is a common phenomenon in Federated Learning.
\vspace{-0.9em}
\section{Conclusion}
\vspace{-0.5em}

We are the first to investigate \prob. Our investigation reveals that existing SSFL/FDG methods underperforms for addressing the challenges in achieving \prob. To address this, we introduce a novel framework, \method, designed to tackle \prob by learning domain invariant features. \method employs a novel alternating two-stage training process. In the first stage, \method trains the server to learn and align features with a parametric distribution, which are then communicated to the clients without incurring any additional communication overhead. In the second stage, \method leverages the server's feature distribution to align client and server features. We conducted extensive experiments on multiple DG datasets and thoroughly evaluated our method, which shows that our \method is the first successful framework capable of achieving \prob. We anticipate our research will highlight \prob and will inspire future research, pushing the boundary of \prob further.
{
    \small
    \bibliographystyle{ieeenat_fullname}
    \bibliography{main}
}

\clearpage
\setcounter{page}{1}
\maketitlesupplementary


\section{Datasets}

We assessed the performance of our proposed \method on five widely used visual benchmarks commonly used for evaluating domain generalization methods. The details of these benchmark datasets are listed below.

\vspace{2mm}



\noindent \textbf{PACS~\cite{yu2022pacs}:} This dataset is a collection of 9,991 images with four distinct domains: art painting, cartoon, photo, and sketch. The task objective is classification across seven classes.

\vspace{2mm}

\noindent \textbf{VLCS~\cite{6751316}:} This dataset comprises 10,729 images spread across four domains, with each domain representing a distinct subdataset. The subdatasets include VOC2007, LabelMe, Caltech-101, and SUN09. The task objective is classification across five different classes.

\vspace{2mm}

 \noindent \textbf{OfficeHome~\cite{venkateswara2017deep}:} OfficeHome dataset is a challenging benchmark composed of four visually distinct domains: Artistic images, Clipart images, Product images, and Real-world images. It comprises 15,500 images distributed across 65 object categories. The task objective is classification across these sixty five classes.
 

\vspace{2mm}

\noindent \textbf{RotatedMNIST~\cite{qiao2023scale}:} This dataset comprises MNIST images~\cite{deng2012mnist} that have been subjected to counter-clockwise rotations at angles of 0, 15, 30, 45, 60, and 75 degrees. These rotations result in six distinct domains: $M_0, M_{15}, M_{30}, M_{45}, M_{60}$, and $M_{75}$. The primary objective remains the classification of ten classes, corresponding to digits 0 through 9. We adopt the dataset variant used in~\cite{fedsr,nguyen2021domain}, where 1,000 images are rotated to define a domain. 

\vspace{2mm}

\noindent \textbf{TerraIncognita~\cite{terra_incognita}:} TerraIncognita dataset is a challenging benchmark composed of four visually distinct domains: L100, L38, L43 and L46. It comprises 24,788 images distributed across 10 classes. The task objective is classification across these ten classes.

\vspace{2mm}

\section{Implementation Details}

For performance evaluation, we allocated one domain as the server dataset and another as the unseen test domain for the final global model, assigning the remaining domains to individual clients. More concretely, in a dataset with $M$ domains, one domain is used for the server, another for testing, and the rest, $M-2$ domains, are distributed among $M-2$ clients. This approach is similar to existing FDG methods~\cite{fedsr,FDG_GA}, where each client possesses data from a unique domain. The accuracy of the final global model is then reported on the unseen test domain. For training, we set the batch size and initial learning rate at 64 and 0.002, respectively. We also set the number of local epochs to 5 and the total communication rounds to 40. After each communication round, client models are averaged using~\cite{fedavg} method skipping the batch normalization parameters as done in~\cite{FedBN}. For optimization, We utilized Stochastic Gradient Descent (SGD) as the optimizer and applied a cosine learning rate decay as the scheduler. The hyperparameters $\alpha$ and $\beta$ are both set to 1, with $\lambda$ set to 0.01 for all experiments. Ablation study of hyper-parameters (e.g., $\alpha$, $\beta$, and $\lambda$) are reported in Ablation section.


\begin{table}[h]
\centering
\setlength{\tabcolsep}{5pt}
\caption{\em Performance comparison of baseline SSFL and \method across two test domains ($M_0, M_{15}$) in the RotatedMNIST dataset. RotatedMNIST dataset consists of six domains: $M_0, M_{15}, M_{30}, M_{45}, M_{60}$ and $M_{75}$. For each combination, we allocated one domain as the server training dataset, another as the unseen test domain for the final global model while assigning the remaining domains to individual clients.}
\scalebox{0.5}{
\begin{tabular}{lc|ccccc|ccccc}
\toprule
\multirow{2}{*}{Method} & \multicolumn{1}{|c|}{Test Domain} & \multicolumn{5}{c|}{\textbf{$M_0$}} & \multicolumn{5}{c}{\textbf{$M_{15}$}} \\ \cline{2-12}
& \multicolumn{1}{|c|}{Server Domain} & $M_{15}$ & $M_{30}$ & $M_{45}$ & $M_{60}$ & $M_{75}$ & $M_0$ & $M_{30}$ & $M_{45}$ & $M_{60}$ & $M_{75}$ \\ \midrule
SSFL & & 77.40 & 56.00 & 37.00 & 23.50 & 15.10 & 68.50 & 67.20 & 41.30 & 38.70 & 36.80 \\ \hline
\rowcolor[gray]{0.95} \method (Ours) & & \textbf{81.90} & \textbf{65.10} & \textbf{55.70} & \textbf{34.90} & \textbf{20.90} & \textbf{84.30} & \textbf{87.00} & \textbf{63.40} & \textbf{52.90} & \textbf{31.00} \\ \bottomrule
\end{tabular}}
\label{tab:rotated_mnist_0_15}
\end{table}


\begin{table}[h]
\centering
\setlength{\tabcolsep}{5pt}
\caption{\em Performance comparison of baseline SSFL and \method across two test domains ($M_{30}, M_{45}$) in the RotatedMNIST dataset. RotatedMNIST dataset consists of six domains: $M_0, M_{15}, M_{30}, M_{45}, M_{60}$ and $M_{75}$. For each combination, we allocated one domain as the server training dataset, another as the unseen test domain for the final global model while assigning the remaining domains to individual clients.}
\scalebox{0.5}{
\begin{tabular}{lc|ccccc|ccccc}
\toprule
\multirow{2}{*}{Method} & \multicolumn{1}{|c|}{Test Domain} & \multicolumn{5}{c|}{\textbf{$M_{30}$}} & \multicolumn{5}{c}{\textbf{$M_{45}$}} \\ \cline{2-12}
& \multicolumn{1}{|c|}{Server Domain} & $M_0$ & $M_{15}$ & $M_{30}$ & $M_{60}$ & $M_{75}$ & $M_0$ & $M_{15}$ & $M_{30}$ & $M_{60}$ & $M_{75}$ \\ \midrule
SSFL & & 43.00 & 71.60 & 75.10 & 54.80 & 41.30 & 32.00 & 50.20 & 76.20 & 70.60 & 53.50 \\ \hline
\rowcolor[gray]{0.95} \method (Ours) & & \textbf{59.40} & \textbf{87.90} & \textbf{84.10} & \textbf{64.20} & \textbf{47.90} & \textbf{47.40} & \textbf{64.70} & \textbf{89.80} & \textbf{88.40} & \textbf{66.30} \\ \bottomrule
\end{tabular}}
\label{tab:rotated_mnist_30_45}
\end{table}

\section{Results on RotatedMNIST}

The evaluation of \method is presented in Tables~\ref{tab:rotated_mnist_0_15},~\ref{tab:rotated_mnist_30_45} and~\ref{tab:rotated_mnist_60_75} on the RotatedMNIST dataset. There are 6 domains in RotatedMNIST dataset: $M_0, M_{15}, M_{30}, M_{45}, M_{60}$, and $M_{75}$. For reporting result of each combination, we allocated one domain as the server training dataset, another as the unseen test domain for the final global model while assigning the remaining domains to individual clients. In RotatedMNIST dataset, we observe a consistent performance improvement with our proposed \method over the baseline SSFL. 


\begin{table}[h]
\centering
\setlength{\tabcolsep}{5pt}
\caption{\em Performance comparison of baseline SSFL and \method across two test domains ($M_{60}, M_{75}$) in the RotatedMNIST dataset. RotatedMNIST dataset consists of six domains: $M_0, M_{15}, M_{30}, M_{45}, M_{60}$ and $M_{75}$. For each combination, we allocated one domain as the server training dataset, another as the unseen test domain for the final global model while assigning the remaining domains to individual clients.}
\scalebox{0.5}{
\begin{tabular}{lc|ccccc|ccccc}
\toprule
\multirow{2}{*}{Method} & \multicolumn{1}{|c|}{Test Domain} & \multicolumn{5}{c|}{\textbf{$M_{60}$}} & \multicolumn{5}{c}{\textbf{$M_{75}$}} \\ \cline{2-12}
& \multicolumn{1}{|c|}{Server Domain} & $M_0$ & $M_{15}$ & $M_{30}$ & $M_{45}$ & $M_{75}$ & $M_0$ & $M_{15}$ & $M_{30}$ & $M_{45}$ & $M_{60}$ \\ \midrule
SSFL & & 23.20 & 40.10 & 54.90 & 76.90 & 78.90 & 21.90 & 28.20 & 37.30 & 46.60 & 77.30 \\ \hline
\rowcolor[gray]{0.95} \method (Ours) & & \textbf{30.00} & \textbf{54.40} & \textbf{73.20} & \textbf{89.40} & \textbf{87.90} & \textbf{24.60} & \textbf{35.90} & \textbf{60.20} & \textbf{65.50} & \textbf{81.60} \\ \bottomrule
\end{tabular}}
\label{tab:rotated_mnist_60_75}
\end{table}


\begin{table}[t]
\centering
\centering\setlength{\tabcolsep}{5pt}
\caption{\em Comparative DG performance of SOTA SSFL and FDG methods and our proposed \method on OfficeHome dataset. The table displays the results across two server training domains: Clipart and Product, with test performance of the global model reported on the unseen Art domain. The GAIN column shows performance improvement of our method compared to the second best method (highlighted by underline).}
\scalebox{0.75}{
\begin{tabular}{lcccc}
\toprule
Method &  Clipart & \color{darkgreen}{GAIN} &  Product & \color{darkgreen}{GAIN} \\
\midrule
CBAFed~\cite{cbafed} & 35.68 & & 30.94 & \\
FedDG~\cite{FedDG} & 28.97 & & 26.82 & \\
FedDG-GA~\cite{FedDG_GA} & 6.39 & & 3.87 & \\
FedGMA~\cite{FedGMA} & 17.72 & & 16.98 & \\
FedSR~\cite{fedsr} & 4.80 & & 5.27 & \\
RScFed~\cite{rscfed} & \underline{40.70} & & \underline{37.37} & \\
SemiFL~\cite{semiFL} & 39.14 & & 36.55 & \\
\rowcolor[gray]{0.95} \method (Ours) & \textbf{48.95} & \color{darkgreen}{+8.25} & \textbf{47.51} & \color{darkgreen}{+10.14} \\
\bottomrule
\end{tabular}}
\label{tab:comparison-with-officehome}
\end{table}


\begin{table}[h]
\centering
\setlength{\tabcolsep}{5pt}
\caption{\em Comparative DG performance of SOTA SSFL and FDG methods and our proposed \method on TerraIncognita dataset. The table displays the results across two server training domains: L38 and L43, with test performance of the global model reported on unseen L100 domain. The GAIN column shows performance improvement of our method compared to the second best method (highlighted by underline).}
\scalebox{0.75}{
\begin{tabular}{lcccc}
\toprule
Method & L38 & \color{darkgreen}{GAIN} & L43 & \color{darkgreen}{GAIN} \\
\midrule
CBAFed~\cite{cbafed} & 35.45 & & \underline{46.25} & \\
FedDG~\cite{FedDG} & 29.62 & & 1.60 & \\
FedDG-GA~\cite{FedDG_GA} & 6.17 & & 27.65 & \\
FedGMA~\cite{FedGMA} & 11.07 & & 8.20 & \\
FedSR~\cite{fedsr} & 8.99 & & 46.22 & \\
RScFed~\cite{rscfed} & 31.56 & & 1.90 & \\
SemiFL~\cite{semiFL} & \underline{36.79} & & 40.58 & \\
\rowcolor[gray]{0.95} \method (Ours) & \textbf{40.17} & \color{darkgreen}{+3.38} & \textbf{48.64} & \color{darkgreen}{+2.39} \\
\bottomrule
\end{tabular}}
\label{tab:comparison-with-terraincognita}
\end{table}

\section{Comparison with SSFL and FDG Methods}
\vspace*{-0.25em}

Here, we compare our proposed \method with SOTA SSFL methods~\cite{semiFL,rscfed,cbafed} as well as SOTA FDG methods~\cite{FedDG,fedsr,FedGMA,FedDG_GA}. We report performance of global model on the unseen Art domain of OfficeHome~\cite{yu2022pacs} dataset in Table~\ref{tab:comparison-with-officehome} and on L100 test domain of TerraIncognita~\cite{terra_incognita} dataset in Table~\ref{tab:comparison-with-terraincognita}. To train using FDG methods, we pretrain the server model with the server dataset and generate pseudo labels prior to client training. The results indicate that the current SSFL methods~\cite{semiFL,rscfed,cbafed} struggles with Domain Generalization (DG). On the other hand, the FDG approaches are incapable of enhancing DG performance, even when trained using pseudo labels since these methods rely heavily on client-labeled data. Nonetheless, whereas existing SSFL and FDG methods struggles, our method thrives on them and successfully generalizes across domains. These results successfully establish the significance of our proposed \method for achieving \prob.

\section{Abltation Study}
All our ablation studies for hyperparameters are conducted using the PACS~\cite{yu2022pacs} benchmark dataset, with Art Painting as the unseen test domain and Cartoon and Photo as the server domains.

\noindent \textbf{Effect of different $\alpha$ \& $\beta$}:
In Table~\ref{tab:alpha_beta}, we present the impact of varying \(\alpha\) and \(\beta\) respectively. From the results, we find that a value of 1 for these parameters delivers optimal results, with any deviation leading to suboptimal performance. The empirical data presented in this table justifies our selection of the hyperparameters \(\alpha\) and \(\beta\).

\noindent \textbf{Effect of $\lambda$}: We report the effect of changing hyperparameter $\lambda$ in Table~\ref{tab:lambda}. The results confirm that a value of $\lambda=0.01$ results in optimal performance. Thus justifying our choice of hyperparameter $\lambda$.

\begin{table}[h]
\centering
\centering\setlength{\tabcolsep}{5pt}
\caption{\em Ablation studies on the effect of jointly changing $\alpha$ and $\beta$ and varying $\lambda$ in the PACS dataset. We report the performance on two server domains, Cartoon and Photo and testing on Art Painting domain, with different values of these hyperparameters.}
\begin{minipage}[h]{0.48\linewidth}
\centering
\scalebox{0.8}{
\begin{tabular}{lcc}
  \toprule
  $\alpha, \beta$ & Cartoon & Photo \\ \midrule
  0.5, 0.5 & 75.49 & 61.52 \\ 
  \rowcolor[gray]{0.95} 1.0, 1.0 & \textbf{75.73} & \textbf{64.40} \\ 
  2.0, 2.0 & 75.34 & 62.84 \\ \bottomrule
\end{tabular}}
\subcaption{Effect of $\alpha$ and $\beta$}
\label{tab:alpha_beta}
\end{minipage}
\hfill
\begin{minipage}[h]{0.48\linewidth}
\centering
\scalebox{0.8}{
\begin{tabular}{lcc}
  \toprule
  $\lambda$ & Cartoon & Photo \\ \midrule
  0.0001 & 68.55 & 57.37 \\ 
  \rowcolor[gray]{0.95} 0.01 & \textbf{75.73} & \textbf{64.40} \\ 
  1.0 & 72.51 & 57.57 \\ \bottomrule
\end{tabular}}
\subcaption{Effect of $\lambda$}
\label{tab:lambda}
\end{minipage}
\end{table}

\noindent \textbf{Effect of Reference matrix}: We report the effect of changing reference matrix $\Sigma$ in Table~\ref{tab:sigma}. We experiment by setting value of $\Sigma = \gamma\Sigmab_k$ by varying $\gamma$ to 50, 100 and 200. We also experiment with minimizing the offdiagonal elements of covariance matrices to 0 without constraining the diagonal elements. The results confirm that diagonal matrix with a value of $\gamma=100$ results in optimal performance. Thus justifying our choice of hyperparameter $\sigma$.

\begin{table}[h]
    \centering
    \setlength{\tabcolsep}{5pt}
    \caption{\em Ablation study on the effect of changing the reference matrix in the PACS dataset. We report the performance on three server domains (Cartoon, Photo, and Sketch) when testing on the Art Painting domain.}
    \scalebox{0.8}{
        \begin{tabular}{llccc}
            \toprule
            $\Lc_{COV}$ & $\gamma$ & Cartoon & Photo & Sketch \\ 
            \midrule
            \multirow{3}{*}{$\frac{1}{m}\| \Sigmab_z - \gamma\Sigmab_k \|^2$} 
            & 0.5 & \textbf{78.32} & 58.98 & 66.94 \\ 
            &  1.0 & 75.73 & \textbf{64.40} & \textbf{67.92} \\ 
            & 2.0 & 68.36 & 51.47 & 63.92 \\ 
            \hline
            $\frac{1}{m}\sum_{i \neq j} [\Sigmab_{z}]_{ij}^2$ & - & 74.17 & 63.04 & 59.96 \\ 
            \bottomrule
        \end{tabular}}
    \label{tab:sigma}
\end{table}


\section{Remaining Results}
\noindent \textbf{PACS:} The evaluation of \method is presented in Table~\ref{tab:pacs_ps} on the PACS dataset. From the results we see that the generalization performance of \method degrades slightly with sketch as test domain. Again this can be attributed to the weaker feature alignment of the remaining domains with sketch domain. Nevertheless, similar to other datasets, we observe a consistent improvement in performance on the Photo test domain with proposed \method compared to baseline SSFL.

\noindent \textbf{VLCS:} The evaluation of \method is presented in Table~\ref{tab:vlcs_ls} on the VLCS dataset. Similarly to other data sets, we observe performance improvement with our proposed \method over the baseline SSFL. 

\noindent \textbf{OfficeHome:} The evaluation of \method is presented in Table~\ref{tab:office-home_pr} on the OfficeHome dataset. Similarly to other data sets, we observe performance improvement with our proposed \method over the baseline SSFL.

\begin{table}[h]
\centering
\setlength{\tabcolsep}{5pt}
\caption{\em Performance comparison of baseline SSFL and \method across different test domains P and S in the PACS dataset. 
}
\scalebox{0.67}{
\begin{tabular}{lc|ccc|ccc}
\toprule
\multirow{2}{*}{Method} & \multicolumn{1}{|c|}{Unseen Test Domain} & \multicolumn{3}{c|}{\textbf{P}} & \multicolumn{3}{c}{\textbf{S}} \\ 
\cline{2-8} 
& \multicolumn{1}{|c|}{Server Trained on} & A & C & S & A & C & P \\ \midrule
SSFL & & 86.05 & 82.93 & 32.34 & \textbf{65.89} & \textbf{75.36} & 30.67 \\ \hline
\rowcolor[gray]{0.95} \method (Ours) & & \textbf{87.31} & \textbf{86.41} & \textbf{79.76} & 64.06 & 71.57 & \textbf{32.50} \\ \bottomrule
\end{tabular}}
\label{tab:pacs_ps}
\end{table}

\begin{table}[h]
\centering\setlength{\tabcolsep}{5pt}
\caption{\em Performance comparison of baseline SSFL and UAP across various test domains L and S within the VLCS dataset~\cite{6751316}. 
}
\scalebox{0.63}{
\begin{tabular}{lc|ccc|ccc}
  \toprule
  \multirow{2}{*}{Method} & \multicolumn{1}{|c|}{Unseen Test Domain}  & \multicolumn{3}{c|}{\textbf{L}} & \multicolumn{3}{c}{\textbf{S}} \\ 
  \cline{2-8} 
                     &  \multicolumn{1}{|c|}{Server Trained on}   & C & S & V & C & L & V \\ \midrule
  SSFL      &          & 46.99      & 54.37 & 56.40   & 44.45      & \textbf{65.39}   & 67.28 \\ 
  \rowcolor[gray]{0.95} UAP (Ours) &         & \textbf{48.49}      & \textbf{58.58} & \textbf{58.73}   & \textbf{50.24}      & 52.86   & \textbf{70.87} \\ \bottomrule
\end{tabular}} 
\label{tab:vlcs_ls}
\end{table}

\begin{table}[h]
\centering\setlength{\tabcolsep}{5pt}
\caption{\em Performance comparison of baseline SSFL and \method across different test domains P and R in the OfficeHome dataset~\cite{venkateswara2017deep}. 
}
\scalebox{0.63}{
\begin{tabular}{lc|ccc|ccc}
  \toprule
  \multirow{2}{*}{Method} & \multicolumn{1}{|c|}{Unseen Test Domain} & \multicolumn{3}{c|}{\textbf{P}}  & \multicolumn{3}{c}{\textbf{R}} \\ 
  \cline{2-8} 
                         & \multicolumn{1}{|c|}{Server Trained on} & A & C & R & A & C & P \\ \midrule 
  SSFL               & & 52.85 & 54.88 & 70.74 & 61.28 & 55.80 & 60.98 \\ 
  \rowcolor[gray]{0.95} \method (Ours)     &     & \textbf{55.96} & \textbf{54.92} & \textbf{72.63} & \textbf{63.92} & \textbf{59.49} & \textbf{64.65} \\ \bottomrule
\end{tabular}}
\label{tab:office-home_pr}
\end{table}


\end{document}